\documentclass[10pt,twocolumn,letterpaper]{article}

\usepackage[pagenumbers]{cvpr} % To force page numbers, e.g. for an arXiv version

\usepackage{multirow}
\usepackage{makecell}
\usepackage{tabularx}
\usepackage[accsupp]{axessibility}

\definecolor{cvprblue}{rgb}{0.21,0.49,0.74}
\usepackage[pagebackref,breaklinks,colorlinks,allcolors=cvprblue]{hyperref}

\usepackage[most]{tcolorbox}
\definecolor{light_green}{HTML}{b2ffb2}
\definecolor{sage}{HTML}{c3efb2}
\definecolor{light_red}{HTML}{ffb2b2}
\definecolor{light_blue}{HTML}{add8e6}
\tcbset{on line, 
        boxsep=1pt, left=1pt,right=1pt,top=1pt,bottom=1pt,
        colframe=white,
        colback=light_green,  
        highlight math style={enhanced}
        }

\usepackage{xcolor}
\usepackage{pifont}
\newcommand{\cmark}{\textcolor{green}{\ding{51}}} % Green check mark
\newcommand{\xmark}{\textcolor{red}{\ding{55}}}

\def\paperID{24} % *** Enter the Paper ID here
\def\confName{CVPR}
\def\confYear{2025}

\title{Classification Drives Geographic Bias in Street Scene Segmentation}

\author{Rahul Nair\\
Arizona State University\\
{\tt\small rnair21@asu.edu}
\and
Bhanu Tokas\\
Arizona State University\\
{\tt\small btokas@asu.edu}
\and
Gabriel Tseng\\
Mila Quebec AI Institute\\
{\tt\small gabrieltseng95@gmail.com}
\and
Esther Rolf\\
University of Colorado Boulder\\
{\tt\small esther.rolf@colorado.edu}
\and
Hannah Kerner\\
Arizona State University\\
{\tt\small hkerner@asu.edu}
}

\begin{document}
\maketitle
\begin{abstract}

Previous studies showed that image datasets lacking geographic diversity can lead to biased performance in models trained on them. Most prior works have studied geo-bias in general-purpose image datasets (e.g., ImageNet, OpenImages) using simple tasks like image classification. Recent works have studied geo-biases in application-based image datasets like driving datasets. However, they have only focused on coarse-grained localization tasks like 2D or 3D detection. In this work, we investigated geo-biases in a Eurocentric driving dataset (Cityscapes) on the fine-grained localization task of instance segmentation. Consistent with previous work, we found that instance segmentation models trained on European driving scenes (Eurocentric models) were geo-biased. Interestingly, we found that geo-biases came from classification errors rather than localization errors, with classification errors alone contributing $10$-$90\%$ of the geo-biases in segmentation and $19$-$88\%$ of the geo-biases in detection. 
Our findings suggest that if a user wants to directly apply region-specific models (e.g., Eurocentric models) globally, they may prefer to coarsen label categories (e.g., use a common label like 4-wheelers over labels like car, bus, and truck). Coarser labels can reduce classification errors, which, as we show in this work, is a major contributor to geo-bias.

\end{abstract}
\section{Introduction}
\label{sec:intro}

\begin{figure*}
    \centering
    \includegraphics[width=0.99\textwidth]{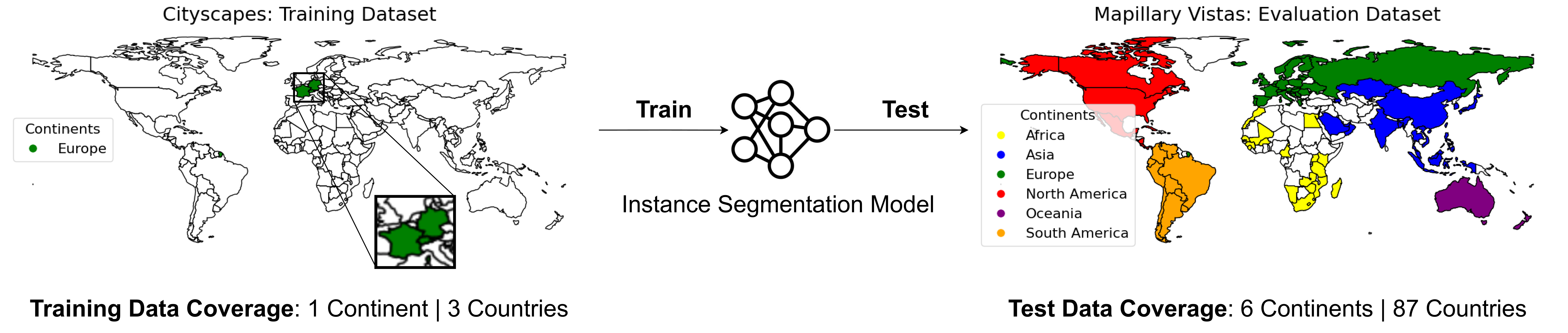}
    \caption{Evaluating Eurocentric models on the Mapillary Vistas dataset}
    \label{fig:teaser_figure}
\end{figure*}

Image datasets like ImageNet \cite{deng2009imagenet} and OpenImages \cite{kuznetsova2020open} have been crucial in advancing computer vision research. However, they have a notable drawback --- they lack geographic diversity. These datasets primarily contain images from Europe and the USA, which do not reflect the diversity of visual scenes worldwide. Previous studies showed that models trained on datasets biased toward some regions (countries or continents) perform well when tested on images from the same region. However, these models perform poorly on underrepresented regions in the dataset \cite{shankar2017no, de2019does}. This variation in model performance across regions, stemming from a lack of geographic diversity in the training dataset, is often referred to as geographic bias or geo-bias.

Identifying (and mitigating) geo-bias is important to avoid the need to retrain models when they encounter images from new geographic locations. Several previous works have studied geo-bias in image datasets. However, these works have predominantly studied general-purpose datasets like ImageNet \cite{deng2009imagenet}, OpenImages \cite{kuznetsova2020open}, and DollarStreet \cite{rojas2022dollar}. These datasets have simple scenes with $1$-$2$ objects of interest per image. Hence, they rarely represent real-world scenarios. Also, these works examined geo-biases in simple tasks like image classification. 

While general-purpose image datasets are popular, they are not the only type of image datasets used in vision research. Some image datasets are also designed for real-world applications --- like driving datasets for autonomous driving. Driving datasets contain images that represent complex real-world street scenes captured from a dashcam. Like general-purpose datasets, well-known driving datasets (e.g., BDD100K \cite{yu2020bdd100k}, Cityscapes \cite{cordts2016cityscapes}) are also collected in Europe and the USA. Since these datasets are geo-biased, it is important to investigate whether models trained on them are also geo-biased. To our knowledge, prior work has only studied geo-bias in driving datasets for coarse-grained localization tasks like 2D and 3D object detection \cite{Wang_2020_CVPR, BiasBehindTheWheel, yang2021st3d}. No prior work has examined geo-bias in driving datasets on the fine-grained localization task of instance segmentation. 

Our study is the first to examine geo-bias in a Eurocentric driving dataset (Cityscapes) for instance segmentation. Cityscapes is a dataset of urban street scenes widely used in computer vision research. All images in Cityscapes are from cities in Germany or neighboring countries. We answer the following research questions: Do vision models trained on Cityscapes perform poorly on continents outside Europe? If there are performance variations among continents (i.e., geo-biases), what is the root cause? We used instance segmentation models pre-trained on Cityscapes, which we call Eurocentric models. We examined geo-bias in these models using out-of-distribution images from Mapillary Vistas (Vistas) \cite{neuhold2017mapillary}, a driving dataset containing images from six continents (refer to Figure \ref{fig:teaser_figure}). 

We define geo-bias as the variation in model performance between continents. We measured geo-bias in Eurocentric models for both instance segmentation and object detection (bounding box regression). We found that Eurocentric models were geo-biased for several classes (rider, bicycle, bus, truck, motorcycle). Interestingly, we found that geo-biases were mainly caused by classification errors --- a phenomenon not identified in previous work. 

We used a new class-merging strategy to isolate the contribution of classification errors to geo-bias. In class-merging, we combine classes that are often misclassified with each other: person-rider (people), car-bus-truck (vehicles with $\geq 4$ wheels), and motorcycle-bicycle ($2$-wheelers). For the geo-biased classes, we found that classification errors alone accounted for $19$-$88\%$ of the geo-biases in detection and $10$-$90\%$ of the geo-biases in segmentation. This revealed that classification is the dominant source of geo-bias in Eurocentric models.

%Our findings show that if a user wants to use region-specific models (like Eurocentric models) globally, it is preferable to use coarser class labels (e.g., $4$-wheeler) rather than fine-grained ones (e.g., car, bus, truck). Coarser classes can reduce classification errors, which, as we show in this work, is a major contributor to geo-bias. In some applications, using coarser classes (over fine-grained ones) may not be feasible. In such cases, we briefly suggest alternate strategies (in Section \ref{sec:discussion}) to counter geo-biases.

In summary, our contributions are as follows: \textbf{(1)} We are the first to examine geo-biases in driving datasets on the fine-grained localization task of instance segmentation; \textbf{(2)} We enable geo-bias analysis in Vistas by providing latitude and longitude information of the images in Vistas; \textbf{(3)} We conduct a systematic analysis of instance segmentation models to identify the root cause of geo-biases; and \textbf{(4)} We propose a new class-merging strategy to measure the isolated contribution of classification errors to geo-bias.

\section{Related Works}
\label{sec:related_work}

Prior studies have examined geo-biases in general-purpose image classification datasets like ImageNet \cite{deng2009imagenet}, OpenImages \cite{kuznetsova2020open}, and DollarStreet \cite{rojas2022dollar}. \citet{shankar2017no} showed that models trained using datasets biased towards North-American or European-centric datasets (ImageNet and OpenImages) had poor performance for countries underrepresented in the training dataset (e.g., Ethiopia and Pakistan) for fine-grained classes (e.g., groom, police officer, vendor). Through experiments with the ImageNet and DollarStreet datasets, \citet{de2019does} showed that models trained using ImageNet, which is dominated by images from high-income countries (Europe and North America), had poorer performance on images from low-income countries for fine-grained classes in DollarStreet (e.g., spices, soap). \citet{rojas2022dollar} showed that training models on DollarStreet, which was more diverse in terms of geography and income than ImageNet, resulted in more balanced performance across geographies and income levels. \citet{gustafson2024exploring} conducted an in-depth analysis of DollarStreet to understand why geo-biases occurred across geographies and income levels. They showed that factors like texture and occlusion contributed to the poor performance of models in low-income continents (e.g., Africa and Asia) in DollarStreet. \citet{ramaswamy2023geode} proposed GeoDE, a geographically diverse image classification dataset created by soliciting images from people across six continents. They showed that models trained on a combined set of ImageNet and GeoDE images performed better on DollarStreet across geographies than models trained solely on ImageNet.

Recently, studies have also explored geo-bias in driving datasets. Chen et al.~\cite{Wang_2020_CVPR} showed that 3D object detection models trained on the Germany-based KITTI~\cite{Geiger2013IJRR} dataset showed a $36\%$ percent drop in performance when tested on images from Waymo~\cite{Sun_2020_CVPR}, a driving dataset collected in the USA. They showed that these performance disparities occurred due to differences in car sizes and shapes between the USA and Europe. Conversely, Yang et al.~\cite{yang2021st3d} showed that models trained on the Waymo dataset showed a greater than $45\%$ decrease in performance when tested on images from the KITTI dataset. Li et al.~\cite{BiasBehindTheWheel} showed that for 2D object detection, models trained on BDD100K, a US-based driving dataset, performed poorly on CityPersons~\cite{Shanshan2017CVPR}, a Germany-based dataset due to difference in skin tones between the countries.

\section{Experiment Design}
\label{sec:exp_design}

\subsection{Instance segmentation task}
We evaluated driving datasets on the instance segmentation task. We chose instance segmentation for several reasons. Instance segmentation is important for scene understanding in autonomous driving use cases. Unlike image classification labels and bounding box annotations (studied in previous work), instance segmentation offers pixel-level localization, which could be important when examining geo-biases. Although semantic segmentation also offers pixel-level localization, it has a limitation. When computing class-IoU, semantic segmentation gives more weight to larger objects in the image. This size bias could mask other important factors causing geo-bias. In contrast, instance segmentation weighs each object equally to compute a class-IoU. This eliminates the size bias problem.

\begin{figure*}
  \centering

  \subfloat[Detection box plots (continent-box-IoUs) for every 
  \{class, model\} pair.]{\includegraphics[width=0.99\textwidth]{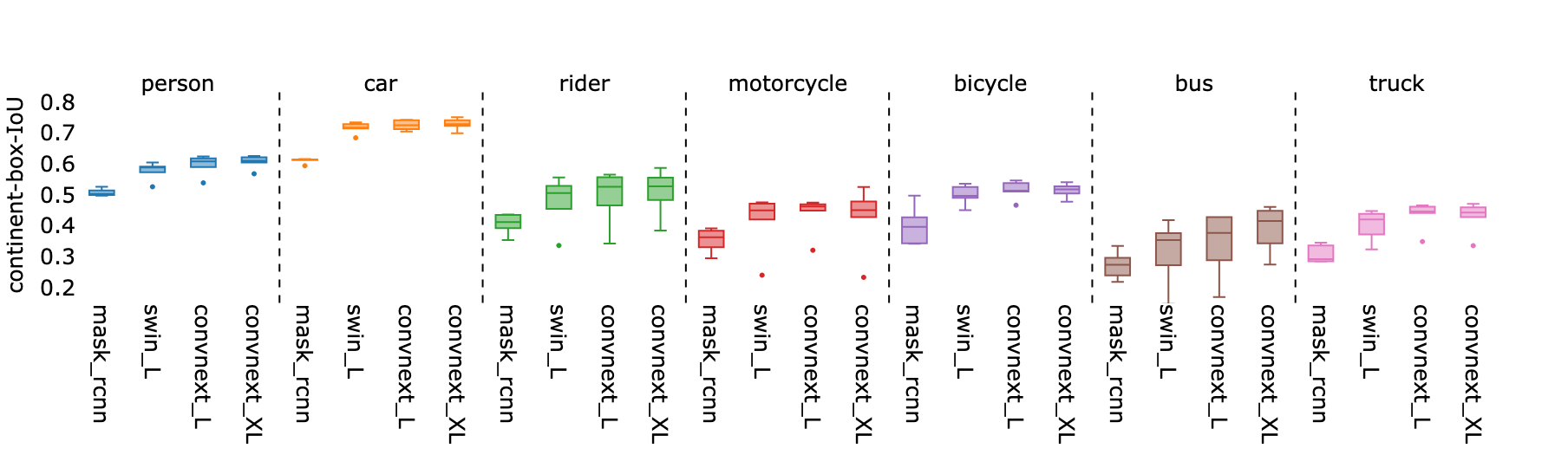}\label{fig:box_plots_bbox_before}}   \vspace{-0.1cm}
  \subfloat[Segmentation box plots (continent-mask-IoUs) for every \{class, model\} pair.]{\includegraphics[width=0.99\textwidth]{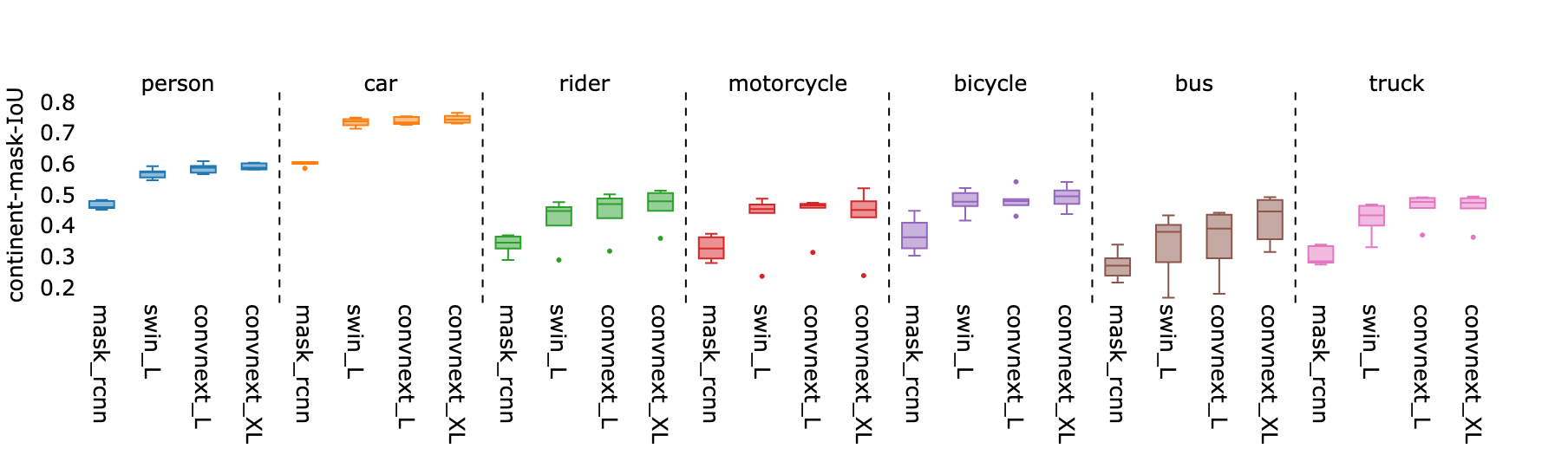}\label{fig:box_plots_seg_before}}
  \caption{Detection and segmentation box plots. Each box plot is made with $6$ data points. Here, a data point refers to the average performance of a model for a class in a continent (continent-IoU). In both figures, while person and car hold minimal geo-biases, other classes (rider, motorcycle, bicycle, bus, and truck) have significant geo-biases.}
  \vspace{-0.4cm}
  \label{fig:box_plots_seg}
\end{figure*}

\begin{figure*}
  \centering

  \subfloat[Detection box plots (corrected-continent-box-IoUs) for every 
  \{class, model\} pair, after class-merging.]{\includegraphics[width=0.99\textwidth]{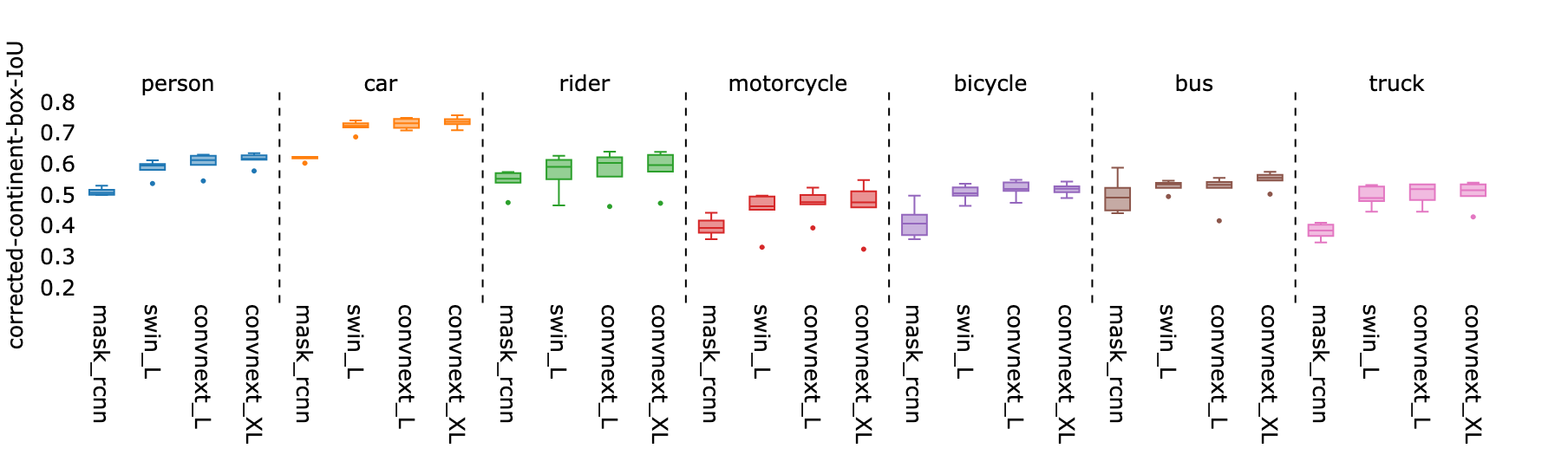}\label{fig:box_plots_bbox_after}} \hfil
  \subfloat[Segmentation box plots (corrected-continent-mask-IoUs) for every \{class, model\} pair, after class-merging.]{\includegraphics[width=0.99\textwidth]{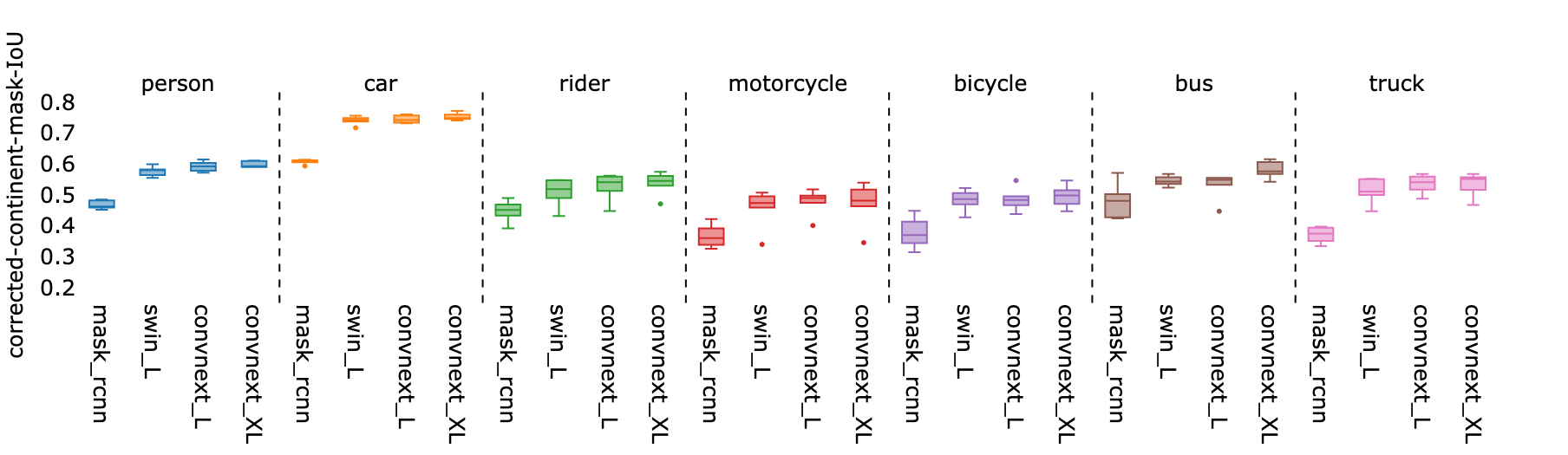}\label{fig:box_plots_seg_after}}
  \vspace{-0.1cm}
  \caption{Detection and segmentation box plots after class-merging. Class-merging was applied to the following groups: car-bus-truck, motorcycle-bicycle, and person-rider. }
  \label{fig:box_plots_bbox}
  \vspace{-0.5cm}
\end{figure*}

\subsection{Models}
We used a total of six models in our experiments to ensure our findings are not limited to a single architecture. We used the Mask-RCNN model \cite{he2017mask} (with ResNet-50 backbone \cite{he2016deep}) since it is widely used for instance segmentation tasks. Given the growing popularity of vision transformers (ViTs), we also evaluated OneFormer \cite{oneformergithub}, which is the current state-of-the-art ViT available for instance segmentation. We used Swin-L, ConvNext-L, and ConvNext-XL backbones for OneFormer. For all models, we examined geo-biases in segmentation performance (using instance masks) and detection performance (using bounding boxes of these instances). Using these models, we covered the two most widely used instance segmentation approaches: detect-and-segment (Mask-RCNN) and segment-only (OneFormer). In addition to the four instance segmentation models, we also used two object detection models: YoloV7 \cite{wang2023yolov7} and Faster-RCNN \cite{ren2016faster}. 

\subsection{Datasets}
\label{subsec:datasets}
All models that we used were pre-trained on the Cityscapes dataset \cite{cordts2016cityscapes}. Cityscapes is a widely used, Eurocentric driving dataset containing complex urban street scenes from cities in Germany and neighboring countries in Europe. The Cityscapes instance segmentation dataset has $3475$ images. The pre-trained model for Mask-RCNN and Faster-RCNN was sourced from MMDetection \cite{chen2019mmdetection}, the OneFormer models were sourced from the author's GitHub repository \cite{oneformergithub}, and the YOLOv7 model was sourced from another Github repository \cite{yolov7_implementation}. 

We used Mapillary Vistas (Vistas) \cite{neuhold2017mapillary} for evaluating models trained on Cityscapes. To our knowledge, Vistas is the only driving dataset with global coverage and instance-level annotations (refer to a comparison of different driving datasets in Section \ref{sec: Dataset_Compare}). As we can see in Table \ref{tab:driving_datasets} of Section \ref{sec: Dataset_Compare}, almost all driving datasets (apart from Vistas \cite{neuhold2017mapillary} and nuScenes \cite{caesar2020nuscenes}) do not provide global coverage as they are collected only in one continent (BDD100K \cite{yu2020bdd100k}, IDD \cite{varma2019idd}, ApolloScapes \cite{huang2018apolloscape}). While nuScenes offers semi-global coverage (Singapore and USA) and instance annotations, it does not label riders separately (nuScenes creates a combined instance mask for motorcycles and riders). The labels in nuScenes are not consistent with Cityscapes (our training dataset). On the other hand, the labels in Vistas are consistent with Cityscapes.

Vistas is an instance segmentation dataset that was originally proposed to improve the geographic diversity of datasets like Cityscapes. Vistas contains $20,000$ (training and validation) images of street scenes on six continents (Europe, North America, South America, Africa, Asia, and Oceania). Since the Vistas dataset does not include geographic metadata, we used the Mapillary API to obtain the longitude and latitude of each image. We removed images for which location metadata could not be obtained. We published the metadata, including Vistas image IDs and their corresponding latitude and longitude information on \url{https://zenodo.org/records/11459554}. 

We combined the training and validation sets into a single evaluation dataset for our experiments. We did not use the Vistas test set because it did not contain location metadata. Our resulting Vistas evaluation dataset contains $10,547$ images (out of the $20,000$ images in Vistas, only $10,547$ images had location metadata). We provide additional information about our Vistas evaluation dataset (number of images per continent, number of instances for each class in a continent) in Section \ref{sec:Vistas_Description}.

We evaluated Eurocentric models trained on Cityscapes using diverse geographic sub-groups of images in Vistas to investigate geo-bias in Eurocentric models. We measured model performance for images from each of the $6$ continents. We could not evaluate at the country level since certain countries (e.g., Uganda, Mozambique) had only one image in Vistas. We chose all classes common to both Cityscapes and Vistas: person, car, bus, truck, bicycle, motorcycle, and rider. Note that Vistas uses two classes --- motorcyclist and bicyclist, instead of one rider class. To ensure consistent labels, we relabeled both classes in Vistas as the rider class. 

We filtered out Vistas images that did not have at any of the classes in our study. Next, we resized these images such that the smaller side of each image was $1024$ pixels. This resizing was done to match the resolution on which the Mask-RCNN was pre-trained. We used the same resolution while evaluating the ViT models (and the detection models) for a fair comparison. Additionally, we excluded point-like objects occupying less than 0.01\% of image pixels, as their lack of shape and texture information limits meaningful insights on geo-biases. 

\subsection{Metrics}

Since we are the first study to examine geo-biases in instance segmentation, we discuss the appropriate metric to compare performance variations across geographies (in our case, continents).

In widely used datasets like Cityscapes and similar benchmarks, the Average Precision (AP) metric is used to evaluate instance segmentation models. AP is calculated across $10$ IoU (intersection-over-union) thresholds, ranging from $0.5$ to $0.95$ in steps of $0.05$. While AP is useful for comparing performance between different models, it may not be suitable for comparing the performance of a single model between different continents.

As an example, consider the performance of a Eurocentric model between two continents: Africa and Asia. Assume the Eurocentric model predicts an IoU in the range $(0, 0.2)$ for all bus instances in Africa and $(0.25, 0.45)$ for all bus instances in Asia. Although there is a disparity in IoUs between Africa and Asia, the bus-AP for both continents would be $0$ (as both precision and recall are $0$), indicating no performance disparity (or geo-bias). This is because the AP metric penalizes bus predictions in both continents due to the high IoU threshold of $0.5$. While lowering the IoU threshold could address this issue, choosing a new threshold introduces human bias.

To eliminate external biases (e.g., human bias) from our geo-bias evaluations, we selected model predictions with the highest IoU (i.e., highest overlap) for the ground truth instances. We then averaged these IoU scores for each class. If the model failed to predict a ground truth instance, we assigned an IoU score of $0$ for that instance.
\section{Results}
\label{sec:results}
\subsection{Are Eurocentric models geo-biased?}
\label{subsec:are_euro_models_biased}

In this study, we analyzed geo-biases in instance segmentation models at two levels of localization: detection and segmentation. We measured geo-biases in segmentation performance (using instance masks) and detection performance (using bounding boxes for these instances). 

First, we calculated the classwise performance of a model in each continent, which we call the continent-IoU:

\begin{equation} \label{eq:continent-IOU}
    %\small
    \text{continent-IOU}_{C,y}(I, \hat{I}) = \frac{1}{N_{C,y}} \sum_{k \in K} IOU(I_{k},\hat{I_{k}})
\end{equation}
Here, $C$ represents a continent (e.g., Europe), and $y$ represents a class (e.g., person). $K$ refers to all instances of class $y$ in continent $C$. $I_{k}$ and $\hat{I_{k}}$ represent the ground-truth and predicted masks for the $k^{th}$ instance. $N_{C,y}$ represents the number of instances of class $y$ in continent $C$ .

Using this formula, we calculated continent-box-IoUs (for detection performance) and continent-mask-IoUs (for segmentation performance) for each \{class, model\} pair. Figure \ref{fig:box_plots_bbox_before} and Figure \ref{fig:box_plots_seg_before} report box plots of continent-box-IoUs and continent-mask-IoUs, respectively. We used box plots because their interquartile ranges (boxes) effectively visualize the geo-biases in a \{class, model\} pair. 

For detection performance, Figure \ref{fig:box_plots_bbox_before} shows that the interquartile range (IQR) for the person (e.g., IQR for Swin-L $= 0.014$) and car (e.g., IQR for Swin-L $= 0.011$) box plots is small, regardless of the model. For the other five classes, we observed significantly larger IQRs for all models. For example, the Swin-L IQR for trucks (IQR $= 0.057$) is nearly $4\times$ that of the person class, and the IQR for buses (IQR $= 0.085$) is nearly $7\times$ that of the person class. 

For segmentation performance, the box plots in Figure \ref{fig:box_plots_seg_before} show a similar trend to those for detection in Figure \ref{fig:box_plots_bbox_before}. Overall, the results in Figures \ref{fig:box_plots_bbox_before} and \ref{fig:box_plots_seg_before} indicate that Eurocentric models have minimal geo-biases for the person and car classes. However, they exhibit significant geo-biases for rider, motorcycle, bicycle, bus, and truck classes.

\subsection{What types of errors are driving geo-bias?}
\label{subsec:what_errors_drive_geo_bias}
\begin{figure*}
    \centering
    \includegraphics[width=0.99\textwidth]{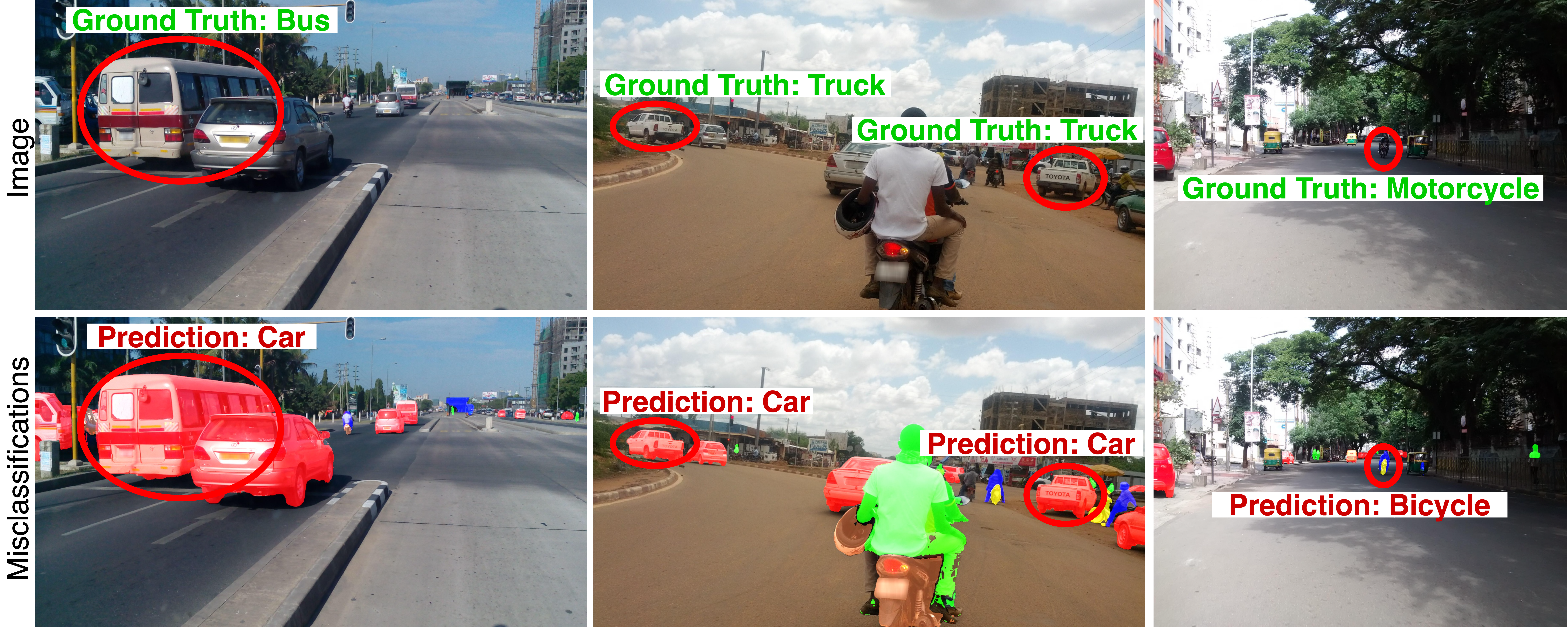}
    \vspace{-0.2cm}
    \caption{Misclassifications by a Eurocentric model (ViT-Swin-L) on images from Africa (left two) and Asia (right one). Objects circled in red were misclassified.}
    \label{fig:geovis_images}
    \vspace{-0.2cm}
\end{figure*}

For the geo-biased classes  (rider, bus, truck, motorcycle, bicycle), Eurocentric models performed the worst in Africa and Asia. Eurocentric models often confused objects in non-European continents (e.g., buses in Africa, motorcycles in Asia) with visually similar classes (e.g., cars, bicycles). In Figure \ref{fig:geovis_images}, we visualized some of the incorrect predictions made by the Swin-L on images from Africa and Asia. 

As shown in the first two columns in Figure \ref{fig:geovis_images}, the Swin-L often misclassified buses and trucks in Africa as cars. 
Mini-buses are a common mode of transportation in Africa but less common in Western countries, where buses are typically larger. These mini-buses resemble large cars or vans in Europe, where they might be annotated with the car class. 

The Swin-L also frequently misclassified motorcycles in Africa and Asia as bicycles, as shown in the third column in Figure \ref{fig:geovis_images}. Many motorcycles in Africa and Asia have thin wheels and are visually similar to bicycles in Europe. 

In a small number of cases, misclassifications occurred due to occlusion. For instance, when a bus was occluded, the model misclassified the bus as a car due to a lack of visual information. Such misclassifications occurred in buses as well as other classes like rider, motorcycle, and truck.

\begin{figure*}
    \centering
    \includegraphics[width=0.99\textwidth]{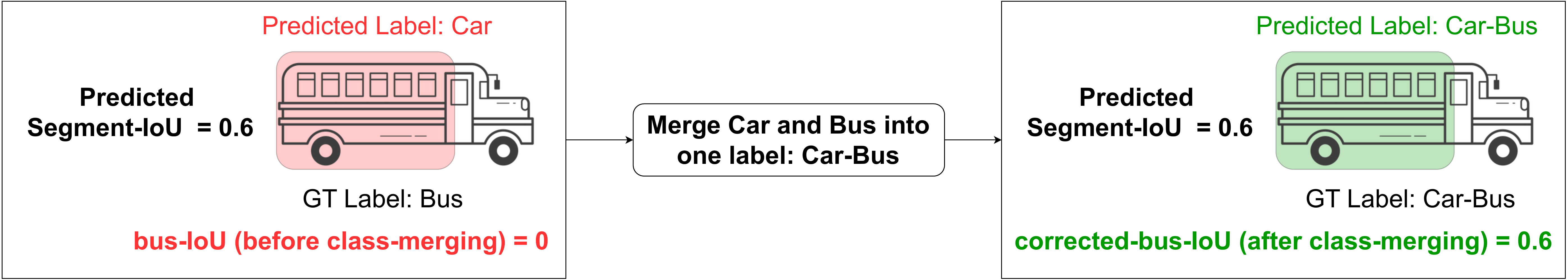}
    \vspace{-0.2cm}
    \caption{An example figure with a bus icon to demonstrate how class-merging resolves misclassification of the bus class and calculates a corrected-bus-IoU.}
    \label{fig:class-merging}
    \vspace{-0.3cm}
\end{figure*}

\begin{table*}

\centering
\caption{Segmentation performance on ViT-Swin-L before (left) and after (right) class-merging. Values on the left of the arrow are continent-mask-IoUs, while values on the right are corrected-continent-mask-IoUs. Values highlighted in green indicate significant performance improvements ($> 0.05$).}
\vspace{-0.15cm}
\label{tab:performance_improvements}
\resizebox{\textwidth}{!}{%
\begin{tabular}{lcccccc}
\toprule
\multicolumn{1}{c}{} & \multicolumn{6}{c}{\textbf{Continents}} \\
\cmidrule(r){2-7}
\textbf{Class} & Europe & Africa & N. America & S. America & Asia & Oceania \\
\midrule
person      & $0.59 \rightarrow 0.60$ & $0.55 \rightarrow 0.56$ & $0.58 \rightarrow 0.58$ & $0.56 \rightarrow 0.56$ & $0.57 \rightarrow 0.58$ & $0.57 \rightarrow 0.58$ \\
car         & $0.75 \rightarrow 0.75$ & $0.71 \rightarrow 0.72$ & $0.74 \rightarrow 0.74$ & $0.74 \rightarrow 0.74$ & $0.73 \rightarrow 0.74$ & $0.75 \rightarrow 0.76$ \\
bus         & \tcbox[colback=light_green]{$0.43 \rightarrow 0.54$} & \tcbox[colback=light_green]{$0.17 \rightarrow 0.52$} & \tcbox[colback=light_green]{$0.28 \rightarrow 0.54$} & \tcbox[colback=light_green]{$0.40 \rightarrow 0.57$} & \tcbox[colback=light_green]{$0.36 \rightarrow 0.55$} & \tcbox[colback=light_green]{$0.40 \rightarrow 0.56$} \\
truck       & \tcbox[colback=light_green]{$0.47 \rightarrow 0.55$} & \tcbox[colback=light_green]{$0.33 \rightarrow 0.45$} & \tcbox[colback=light_green]{$0.43 \rightarrow 0.50$} & \tcbox[colback=light_green]{$0.40 \rightarrow 0.51$} & \tcbox[colback=light_green]{$0.47 \rightarrow 0.55$} & \tcbox[colback=light_green]{$0.44 \rightarrow 0.51$} \\
bicycle     & $0.46 \rightarrow 0.47$ & $0.51 \rightarrow 0.51$ & $0.49 \rightarrow 0.50$ & $0.52 \rightarrow 0.52$ & $0.42 \rightarrow 0.43$ & $0.46 \rightarrow 0.47$ \\
motorcycle  & $0.46 \rightarrow 0.46$ & \tcbox[colback=light_green]{$0.24 \rightarrow 0.34$} & $0.47 \rightarrow 0.50$ & $0.44 \rightarrow 0.47$ & $0.45 \rightarrow 0.48$ & $0.49 \rightarrow 0.51$ \\
rider       & \tcbox[colback=light_green]{$0.46 \rightarrow 0.55$} & \tcbox[colback=light_green]{$0.29 \rightarrow 0.43$} & \tcbox[colback=light_green]{$0.48 \rightarrow 0.55$} & \tcbox[colback=light_green]{$0.44 \rightarrow 0.51$} & \tcbox[colback=light_green]{$0.40 \rightarrow 0.49$} & \tcbox[colback=light_green]{$0.46 \rightarrow 0.52$} \\
\bottomrule
\end{tabular}%
}
\vspace{-0.35cm}
\end{table*}

\subsection{What is the contribution of classification errors in geo-bias?}
\label{subsec:classification_contribution}
To measure the isolated contribution of classification errors on geo-bias, we used a class-merging strategy. In class-merging, we merged classes that were visually similar (and often misclassified with each other) during evaluation. This allowed us to recalculate the IoU of an object as if it had been correctly classified. 

%Since misclassifications were common in non-European continents, we measured their isolated contribution to geo-bias using a class-merging strategy.
 
Assume that the bus icon shown in Figure \ref{fig:class-merging} is originally labeled as a bus. Assume an instance segmentation model predicted a segment that has an IoU of $0.6$ with this bus, but the model mistakenly classified the bus as a car. Hence, the IoU of the bus class is $0$. To recalculate the IoU of the bus as if it had been correctly classified, we modified the ground-truth class of the bus to car-bus. At the prediction stage, all car and bus predictions were also relabeled as car-bus. The bus that was previously predicted as a car is now predicted as the car-bus class. Since the ground-truth label matches the predicted label (car-bus), the model will return an IoU value of $0.6$ for car-bus. We map this car-bus IoU value of $0.6$ back to the original bus class. We refer to this updated IoU as the corrected-IoU of the bus class.

We created the following groups for class-merging: car-bus-truck, motorcycle-bicycle and person-rider. These groupings were based on the general structure of objects. For instance, car-bus-truck contains $4$-wheeled vehicles, motorcycle-bicycle contains $2$-wheeled vehicles, and person-rider contains humans. We applied class-merging to the detection and segmentation outputs and generated corrected-continent-box-IoUs and corrected-continent-mask-IoUs, respectively, for each class. 

In Table \ref{tab:performance_improvements}, for each class, we showed the segmentation performance before and after class-merging for the Swin-L. The continent-mask-IoUs are shown on the left side of the arrow, while the corrected-continent-mask-IoUs are shown on the right side. We see a significant improvement in IoUs after resolving misclassifications using class-merging (especially in Africa). This improvement in IoU shows the substantial impact of misclassifications on the Swin-L's poor performance in non-European continents.

In Table \ref{tab:performance_improvements}, we see that the performance of Swin-L also improved in Europe for classes like bus, truck, and rider. This shows that Eurocentric models made misclassifications in Europe as well. As discussed earlier, in some cases, misclassifications occurred when objects were occluded. It is these occlusion-based misclassifications that were resolved using class-merging, which explains why the performance in Europe improved after class-merging for some classes.

To evaluate the impact of misclassifications on geo-bias, we examined changes in box plots (IQRs) before and after class-merging. Compared to Figures \ref{fig:box_plots_bbox_before} and \ref{fig:box_plots_seg_before} (before class-merging), the IQR of the box plots in Figures \ref{fig:box_plots_bbox_after} and \ref{fig:box_plots_seg_after} (after class-merging) significantly reduced for the geo-biased classes (bus, motorcycle, bicycle, rider, and truck). %This notable reduction in IQR indicates that misclassifications are a prominent source of geo-bias.

While class-merging reduced IQR for several classes (as IoU improved across continents), it does not mean that class-merging mitigated geo-bias. The lower IQR only indicates that misclassifications are a major source of geo-bias. Next, we used class-merging to measure the percentage contribution of misclassifications on geo-bias.

%While class-merging reduced IQR (as IoU improved across continents), this does not mean that class-merging mitigated geo-bias. This reduction in IQR only indicates that misclassification is a prominent source of geo-bias, and class-merging can be used to measure this contribution.

%To measure the contribution of misclassifications on geo-bias,

We quantified geo-bias using the Coefficient of Variation (CV). CV measures the variation in model performance across continents, consistent with our definition of geo-bias in Section \ref{sec:intro}. CV is  the ratio of the standard deviation ($\sigma$) of continent-IoUs to the mean continent-IoU for a \{class, model\} pair: $\text{CV} = \sigma_{\text{continents}} / \mu_{\text{continents}}$. We used CV instead of standard deviation because CV's normalization makes it scale invariant. 

To measure the contribution of classification errors in detection, we computed two measures: $\text{CV}_{\text{det}}$, the geo-bias in detection performance originally (before class-merging), and $\text{CV}_{\text{det-corrected}}$, the geo-bias in detection performance after class-merging. We then computed the $\%$ change in geo-bias before and after class-merging:

\begin{equation} \label{eq:det_disparity}
\Delta\text{CV}_{\text{det-det-corrected}} = \frac{\text{CV}_{\text{det-corrected}} - \text{CV}_{\text{det}}}{\text{CV}_{\text{det}}} \times 100    
\end{equation}

In the second column of Table \ref{tab:disparity}, for each class, we show the $\Delta\text{CV}_{\text{det-det-corrected}}$ values for the Swin-L model. The percentage reduction in CV after class-merging is significant for all geo-biased classes: $19\%$ for bicycles, $36\%$ for motorcycles, $37\%$ for riders, $45\%$ for trucks, and $88\%$ for buses. This shows that misclassifications contributed significantly to the geo-biases in detection performance. There were small changes in CV for the person and car classes. This is because, for person and car, the performance variations among continents were already negligible (evident from the box-plots before class merging, in Figure \ref{fig:box_plots_bbox_before}).

Similar to detection, we measured the contribution of misclassifications on geo-bias in segmentation using the corresponding CV measures: 

\begin{equation} \label{eq:seg_disparity}
\Delta\text{CV}_{\text{seg-seg-corrected}} = \frac{\text{CV}_{\text{seg-corrected}} - \text{CV}_{\text{seg}}}{\text{CV}_{\text{seg}}} \times 100    
\end{equation}

In the third column of Table \ref{tab:disparity}, for each class, we show the $\Delta\text{CV}_{\text{seg-seg-corrected}}$ values for the Swin-L model. Similar to detection, the reduction in CV for segmentation is significant for all geo-biased classes: $10\%$ for bicycles, $37\%$ for trucks, $39\%$ for motorcycles, $47\%$ for riders, and $90\%$ for buses. As expected, there were minimal changes in CV for the person and car classes. This shows that misclassifications were a prominent source of geo-bias in segmentation.  

We observed similar reductions in CV after class-merging for all the other instance segmentation models: Mask-RCNN, ConvNext-L, and ConvNext-XL. We discuss the $\Delta\text{CV}_{\text{det-det-corrected}}$ and the $\Delta\text{CV}_{\text{seg-seg-corrected}}$ values for these models in Section \ref{sec: Extra_Models}. In addition to instance segmentation models, we show that misclassifications are a prominent source of geo-bias in detection models. We discuss the $\Delta\text{CV}_{\text{det-det-corrected}}$ values for two detection models: YOLOV7 and Faster-RCNN in Section \ref{sec: Extra_Models}.

In conclusion, misclassifications significantly contributed to geo-bias in instance segmentation (and detection) models. For the geo-biased classes (bus, truck, bicycle, motorcycle, and rider), misclassifications accounted for $19$-$88\%$ of the geo-biases in detection and $10$-$90\%$ of the geo-biases in segmentation. In contrast, classes like person and car showed minimal $\Delta$CV values, indicating no significant geo-bias before class merging, and misclassifications were not a major source of error for these classes.

\begin{table}
\centering
\caption{Percentage change in CV for the Swin-L. Column $2$ shows the $\%$ change in CV in detection performance before and after class-merging ($\Delta\text{CV}_{\text{det-det-corrected}}$). Column $3$ shows the $\%$ change in CV in segmentation performance before and after class-merging ($\Delta\text{CV}_{\text{seg-seg-corrected}}$).}
\label{tab:disparity}
\resizebox{0.45\textwidth}{!}{
\begin{tabular}{lcc}
\toprule
Class & $\Delta\text{CV}_{\text{det-det-corrected}}$ & $\Delta\text{CV}_{\text{seg-seg-corrected}}$ \\
\midrule
person  & $-5.78\%$ & $-5.32\%$ \\ 
car     & $6.07\%$ & $0.61\%$  \\
bus     & $-88.68\%$ & $-90.07\%$  \\
truck     & $-45.14\%$ & $-37.59\%$   \\
bicycle & $-19.44\%$ & $-10.62\%$   \\
motorcycle & $-36.29\%$ & $-39.66\%$  \\
rider   & $-37.38\%$ & $-47.44\%$  \\
\bottomrule
\end{tabular}
}
\end{table}

\section{Discussion}
\label{sec:discussion}

\subsection{Applying geo-biased models globally}

Consistent with previous works, we show that instance segmentation models are geo-biased when trained on geographically-biased datasets (e.g., Cityscapes). We showed that in Eurocentric instance segmentation models, classification errors are a prominent source of geo-bias.

Since region-specific models (e.g., Eurocentric models) are geo-biased, they should be adapted for new geographic regions before being deployed in those regions. For instance, if we want to use Eurocentric models in Africa, these models could be fine-tuned on some street scenes from Africa. Future work could investigate bias mitigation strategies such as training using geography-aware classification techniques to reduce classification errors in new regions \cite{mac2019presence, Wang_2020_CVPR}. Our results also emphasize the need for larger, globally representative driving datasets that can be used to train models with better global applicability.

If a user intends to deploy region-specific models directly to new regions, our results suggest that it would be beneficial to use coarser labels rather than fine-grained labels (e.g., four-wheeler instead of car, bus, and truck or two-wheeler instead of motorcycle and bicycle). Coarser classes result in fewer classification errors, which are the main contributor to geo-bias in Eurocentric models. Thus, the decrease in label specificity may 
be worth gains to geographic domain adaptation in some cases. 

\subsection{Limitations}

Our study is the first to provide insights into geographic biases in driving datasets for 2D instance segmentation. We evaluated Eurocentric models (trained on Cityscapes) using Mapillary Vistas, which has global coverage. We could not test on additional datasets since Vistas is the only publicly available 2D driving dataset that offers global coverage and instance-level annotations, as shown in Section \ref{sec: Dataset_Compare}.

We used models pre-trained only on the Eurocentric Cityscapes dataset as, pre-trained model weights were unavailable for other geo-biased driving datasets with instance segmentation annotations (Section \ref{sec: Dataset_Compare}). Future work could investigate geo-biases arising from models trained on other geo-biased datasets (e.g., Americentric datasets like BDD100k \cite{yu2020bdd100k}) by training model architectures on these datasets from scratch. 

In our study, we examined seven classes that were common to both Cityscapes and Vistas. Most datasets (including Cityscapes and Vistas) typically provide instance-level annotations only for objects critical to driving safety, such as people, cars, and trucks. Several important classes that could have added value to this study (e.g., traffic signs and traffic lights) did not have instance-level annotations and thus could not be analyzed in our study.

We also point out a limitation in our class-merging strategy. In class-merging, we manually identified classes that were visually similar and merged them during evaluation. Future research could explore ways to automate class-merging, perhaps using metrics that measure visual similarity among classes.

Although we provided country-level metadata for the Vistas images (as shown in Section \ref{subsec:datasets}), we examined geo-biases at the continent level. This is because some countries had very few images in the dataset. For instance, countries like Uganda and Mozambique (in Africa) had only one image each in Vistas. Since we examined geo-biases at the continent level, we could not capture performance variations between countries.

\section{Conclusion}

We investigated geo-bias in instance segmentation models trained on Cityscapes, a Eurocentric driving dataset. We used Vistas, a driving dataset with global coverage, to test the performance of Eurocentric models on driving scenes from new geographic regions. 

We observed that geo-biases were present for some classes (bus, truck, motorcycle, bicycle, rider) but not for others (person, car). We showed that these geo-biases were caused primarily by classification errors. Eurocentric models often misclassified objects in non-European continents (e.g., buses in Africa, motorcycles in Asia) with visually similar objects in Europe (e.g., cars, bicycles). 

We used a class-merging strategy to isolate the contribution of classification errors. We showed that for geo-biased classes, classification errors contributed to $19$-$88\%$ of the geo-biases in detection and $10$-$90\%$ of the geo-biases in segmentation. This suggests that geo-bias is largely caused by classification errors, not localization errors. 

%To address classification errors, in class-merging, we grouped fine-grained classes into simpler, coarser classes (e.g., four-wheelers over car vs. bus vs. truck, two-wheelers over motorcycle vs. bicycle, and humans over person vs. rider). Our findings suggest that if a user wants to directly apply region-specific models (e.g., Eurocentric models) to new regions (without adapting models for these regions), it would be beneficial to use coarser class definitions. As shown through class-merging, coarser classes can reduce misclassifications, which is a prominent source of geo-bias in Eurocentric models.

%In applications where fine-grained classes are essential, alternate strategies like training and testing models within the same geographic region or geography-aware classification should be used to reduce geo-biases.

%Our findings show that if we wish to use region-specific models (Eurocentric in our case) globally, coarser classes can avoid geo-biases arising from model confusions, which is a significant percentage. In applications where fine-grained classes are essential, alternate strategies like train and test on the same geographic region or geographiary aware classification should be used to reduce geo-biases.
{
    \small
    \bibliographystyle{ieeenat_fullname}
    \bibliography{main}
}

% WARNING: do not forget to delete the supplementary pages from your submission 
\clearpage
\setcounter{page}{1}
\maketitlesupplementary

\appendix

\section{Dataset Comparision}
\label{sec: Dataset_Compare}

We compared some of the widely used driving datasets in Table \ref{tab:driving_datasets}. It is evident that Mapillary Vistas (Vistas) \cite{neuhold2017mapillary} is the only dataset with global coverage and instance-level annotations.
\begin{comment}
\begin{table*}
\centering
\caption{A comparison of well-known driving datasets.}
\label{tab:driving_datasets}
\resizebox{\textwidth}{!}{
\begin{tabular}{lcc}
\toprule
Driving Datasets & Continents Covered & Instance Segmentation \\
\midrule
IDD \cite{varma2019idd}  & Asia (India) & \xmark \\
BDD100K \cite{yu2020bdd100k} & N. America (USA) & \cmark \\ 
Cityscapes \cite{cordts2016cityscapes}  & Europe & \cmark \\ 
nuScenes \cite{caesar2020nuscenes}  & N. America (USA) and Asia (Singapore) & \cmark \\ 
ApolloScapes \cite{huang2018apolloscape}  & Asia (China) & \cmark \\ 
Mapillary Vistas \cite{neuhold2017mapillary}   & N. America, S. America, Europe, Africa, Asia, and Oceania & \cmark \\ 
\bottomrule
\end{tabular}
}
\end{table*}
\end{comment}

\section{Additional Details on Vistas}
\label{sec:Vistas_Description}

In Table \ref{tab:img_per_continent}, we show the number of images in each continent in Vistas. Table \ref{tab:class_distribution} shows the number of instances of each class in each continent. Tables \ref{tab:img_per_continent} and \ref{tab:class_distribution} confirm that we had sufficient instances across continents (for most classes) to conduct robust geo-bias evaluations.

%To compute Table \ref{tab:class_distribution}, we used the $10,547$ images obtained after the pre-processing stage

%Note that the total count of images in Table \ref{tab:img_per_continent} ($10,547$) is lesser than the number of images in our Vistas evaluation set ($10,900$ reported in Section \ref{subsec:datasets}). This is because a few images were eliminated in pre-processing (pre-processing steps are discussed in Section \ref{subsec:datasets}). 

\section{Class-merging on additional models}
\label{sec: Extra_Models}

In addition to Swin-L (shown in Table \ref{tab:disparity}), we applied class-merging on the following instance segmentation models: Mask-RCNN, ConvNext-L, and ConvNext-XL. Table \ref{tab:disparity_appendix_ins_seg} shows the $\Delta\text{CV}_{\text{det-det-corrected}}$ and the $\Delta\text{CV}_{\text{seg-seg-corrected}}$ values for these models. Similar to Swin-L, for the geo-biased classes, the CV values for detection and segmentation reduced for all models (Mask-RCNN, ConvNext-L, ConvNext-XL) after class-merging. This shows that classification errors significantly contributed to geo-biases in instance segmentation models.

To measure the impact of classification errors on geo-biases in detection models, we also applied class-merging to well-known object detection models (trained on Cityscapes): YOLOv7 \cite{wang2023yolov7} and Faster-RCNN \cite{ren2016faster}. Table \ref{tab:disparity_appendix_det} shows the $\Delta\text{CV}_{\text{det-det-corrected}}$ values for these models. Similar to the segmentation models (shown in Tables \ref{tab:disparity} and \ref{tab:disparity_appendix_ins_seg}), the CV values for detection reduced for YOLOv7 and Faster-RCNN after class-merging. 

Interestingly, in Faster-RCNN, the CV increased for buses ($32\%$) and slightly for motorcycles ($9\%$). To explain the increase in CV, we show the continent-box-IoUs for buses and motorcycles before and after class-merging in Table \ref{tab:perf_improvement_bus}. 

For buses, the detection performance of Faster-RCNN before class-merging was similar across continents. After class-merging, performance improved significantly in non-European continents like Africa ($0.36$ to $0.61$), North America ($0.30$ to $0.51$), and Oceania ($0.37$ to $0.54$), compared to a smaller improvement in Europe ($0.30$ to $0.43$). This led to an increase in CV. Similarly, for motorcycles, non-European continents like North America ($0.40$ to $0.45$), South America ($0.39$ to $0.48$), and Asia ($0.42$ to $0.46$) saw larger gains than Europe ($0.40$ to $0.42$). 

Despite the increased CV in buses (and marginally in motorcycles) for Faster-RCNN, the results show that classification errors significantly impacted performance in non-European continents. 

\begin{table*}[ht]
\centering
\caption{A comparison of well-known driving datasets.}
\label{tab:driving_datasets}
\resizebox{\textwidth}{!}{
\begin{tabular}{lccccccc}
\toprule
Driving Datasets & \multicolumn{6}{c}{Continents Covered} & Instance Segmentation \\
 & Asia & Europe & N. America & S. America & Africa & Oceania & \\
\midrule
IDD \cite{varma2019idd}  & \cmark & \xmark & \xmark & \xmark & \xmark & \xmark & \xmark \\
BDD100K \cite{yu2020bdd100k} & \xmark  & \xmark & \cmark & \xmark & \xmark & \xmark & \cmark \\ 
Cityscapes \cite{cordts2016cityscapes}  & \xmark & \cmark  & \xmark & \xmark & \xmark & \xmark & \cmark \\ 
nuScenes \cite{caesar2020nuscenes}  & \cmark  & \xmark & \cmark & \xmark & \xmark & \xmark & \cmark \\ 
ApolloScapes \cite{huang2018apolloscape}  & \cmark & \xmark & \xmark  & \xmark & \xmark & \xmark & \xmark \\ 
Mapillary Vistas \cite{neuhold2017mapillary}  & \cmark & \cmark & \cmark & \cmark & \cmark & \cmark & \cmark \\ 
\bottomrule
\end{tabular}
}
\end{table*}

\begin{table*}[ht]
\centering
\caption{Number of images per continent in the Vistas dataset.}
\begin{tabular}{cccccccc}
\toprule
Continent & Asia & Europe & N. America & S. America & Africa & Oceania & Total \\
\midrule
Number of Images & 2157 & 3752 & 3090 & 712 & 154 & 682 & 10547\\
\bottomrule
\end{tabular}
\label{tab:img_per_continent}
\end{table*}

\begin{table*}[t]
    \centering
    \caption{Number of instances of each class in each continent in the Vistas dataset.}
    \begin{tabular}{lcccccc}
        \toprule
         Class & Asia & Europe & N. America & S. America & Africa & Oceania\\
         \midrule
         person & 1892 & 10106 & 4265 & 2197 & 833 & 1109 \\
         car & 9514 & 21848 & 22068 & 4820 & 1090 & 4256 \\
         bus & 459 & 943 & 435 & 350 & 80 & 94 \\
         truck & 1205 & 886 & 1057 & 275 & 84 & 165 \\
         bicycle & 411 & 1691 & 380 & 160 & 24 & 69 \\
         motorcycle & 743 & 1022 & 149 & 388 & 128 & 37 \\
         rider & 771 & 1178 & 322 & 469 & 127 & 62 \\
        \bottomrule
    \end{tabular}
    \label{tab:class_distribution}
\end{table*}

\begin{table*}[t]
\centering
\caption{Detection performance on Faster-RCNN for buses and motorcycles before and after class-merging. Values on the left of the arrow are continent-box-IoUs, while values on the right are corrected-continent-box-IoUs.}
\label{tab:perf_improvement_bus}
\resizebox{\textwidth}{!}{%
\begin{tabular}{lcccccc}
\toprule
\multicolumn{1}{c}{} & \multicolumn{6}{c}{\textbf{Continents}} \\
\cmidrule(r){2-7}
\textbf{Class} & Europe & Africa & N. America & S. America & Asia & Oceania \\
\midrule
bus  & $0.30 \rightarrow 0.43$ & $0.36 \rightarrow 0.61$ & $0.30 \rightarrow 0.51$ & $0.33 \rightarrow 0.49$ & $0.29 \rightarrow 0.44$ & $0.37 \rightarrow 0.54$ \\
motorcycle  & $0.40 \rightarrow 0.42$ & $0.36 \rightarrow 0.43$ & $0.40 \rightarrow 0.45$ & $0.39 \rightarrow 0.48$ & $0.42 \rightarrow 0.46$ & $0.37 \rightarrow 0.40$ \\
\bottomrule
\end{tabular}%
}
\end{table*}

\begin{table}[b!]
\centering
\caption{Percentage change in CV for various instance segmentation models. Column $2$ shows the $\%$ change in CV in detection performance before and after class-merging ($\Delta\text{CV}_{\text{det-det-corrected}}$). Column $3$ shows the $\%$ change in CV in segmentation performance before and after class-merging ($\Delta\text{CV}_{\text{seg-seg-corrected}}$).}
\label{tab:disparity_appendix_ins_seg}
%\resizebox{0.45\textwidth}{!}{
    
\small
\begin{tabular}{lccc}
\toprule
Model Name & Class & $\Delta\text{CV}_{\text{det-det-corrected}}$ & $\Delta\text{CV}_{\text{seg-seg-corrected}}$ \\
\midrule
Mask-RCNN & person      & $4.05\%$ & $2.72\%$ \\
          & car         & $-5.97\%$  & $-6.97\%$  \\
          & bus         & $-29.06\%$ & $-29.18\%$ \\
          & truck       & $-28.84\%$ & $-24.96\%$ \\
          & bicycle     & $-15.41\%$ & $-11.45\%$ \\
          & motorcycle  & $-26.83\%$ & $-17.03\%$ \\
          & rider       & $-15.64\%$ & $-19.35\%$ \\
\midrule
ConvNext-L & person     & $-0.65\%$ & $1.34\%$ \\
          & car         & $4.87\%$  & $1.44\%$  \\
          & bus         & $-66.61\%$ & $-72.34\%$ \\
          & truck       & $-29.97\%$ & $-42.22\%$ \\
          & bicycle     & $-7.99\%$ & $-1.10\%$  \\
          & motorcycle  & $-30.42\%$ & $-40.02\%$ \\
          & rider       & $-34.12\%$ & $-46.19\%$ \\
\midrule
ConvNext-XL & person    & $-2.76\%$ & $-0.23\%$ \\
          & car         & $-10.43\%$  & $-14.04\%$  \\
          & bus         & $-74.10\%$ & $-72.78\%$ \\
          & truck       & $-29.14\%$ & $-34.48\%$ \\
          & bicycle     & $-16.60\%$ & $-3.39\%$  \\
          & motorcycle  & $-30.96\%$ & $-37.36\%$ \\
          & rider       & $-27.12\%$ & $-44.82\%$ \\
\bottomrule
\end{tabular}
%}
\end{table}

\newpage

\begin{table}[t!]
\centering
\caption{Percentage change in CV for various detection models. Column $2$ shows the $\%$ change in CV in detection performance before and after class-merging ($\Delta\text{CV}_{\text{det-det-corrected}}$)}.
\label{tab:disparity_appendix_det}
%\resizebox{0.45\textwidth}{!}{
\small
\begin{tabular}{ccc}
\toprule
Model Name & Class & $\Delta\text{CV}_{\text{det-det-corrected}}$ \\
\midrule
YOLOv7    & person      & $-5.81\%$  \\
          & car         & $2.59\%$    \\
          & bus         & $-55.49\%$  \\
          & truck       & $-34.21\%$  \\
          & bicycle     & $-0.47\%$   \\
          & motorcycle  & $-33.13\%$  \\
          & rider       & $-14.23\%$  \\
\midrule
Faster-RCNN & person      & $-2.66\%$  \\
          & car         & $-6.46\%$  \\
          & bus         & $32.55\%$  \\
          & truck       & $-51.98\%$  \\
          & bicycle     & $-1.85\%$  \\
          & motorcycle  & $9.41\%$ \\
          & rider       & $-6.79\%$ \\
\bottomrule
\end{tabular}
%}
\end{table}

\end{document}